\documentclass[10pt,twocolumn]{article}

\usepackage[margin=1in]{geometry}
\usepackage[T1]{fontenc}
\usepackage[utf8]{inputenc}
\usepackage{microtype}
\usepackage{hyperref}
\usepackage{url}
\usepackage{booktabs}
\usepackage{graphicx}
\usepackage{enumitem}
\usepackage{authblk}

\usepackage{biblatex}      % For bibliography handling
\usepackage{listings}
\begin{document}

\title{A Baseline Multimodal Approach to Emotion Recognition in Conversations}

\author[1]{V\'ictor Yeste\thanks{Corresponding author: victor.yeste@universidadeuropea.es.}} 
\author[1]{Rodrigo Rivas-Ar\'evalo}
\affil[1]{School of Science, Engineering and Design, Universidad Europea de Valencia, 46010 Valencia, Spain}

\date{}

\maketitle

\begin{abstract}
We present a lightweight multimodal baseline for emotion recognition in conversations using the SemEval-2024 Task 3 dataset built from the sitcom \textit{Friends}. The goal of this report is not to propose a novel state-of-the-art method, but to document an accessible reference implementation that combines (i) a transformer-based text classifier and (ii) a self-supervised speech representation model, with a simple late-fusion ensemble. We report the baseline setup and empirical results obtained under a limited training protocol, highlighting when multimodal fusion improves over unimodal models. This preprint is provided for transparency and to support future, more rigorous comparisons.
\end{abstract}

\begin{keywords}
  Emotion Recognition in Conversations,
  Multimodal Baseline,
  SemEval-2024 Task 3,
  Transformers,
  Speech Representation Learning.
\end{keywords}

\vspace{0.5em}
\noindent\textbf{Disclaimer (baseline / technical report).} This manuscript is a baseline study derived from an undergraduate final-year project (TFG) and an internal replication effort. It is shared as a technical report/preprint for transparency and reproducibility. The system is not intended to be state-of-the-art, and the experimental protocol is intentionally lightweight (limited hyperparameter tuning and limited evaluation). Readers should interpret the results as a reference point rather than a definitive contribution.

\noindent Code and materials are available at the accompanying repository.\footnote{\url{https://github.com/RodrigoRivasArevalo/TFG_Multimodal_final}}

\vspace{0.75em}

\section{Introduction}

Emotion recognition in conversations is an active area of affective computing, with applications ranging from human--computer interaction to analysis of social dialogue. This technical report documents a practical multimodal \emph{baseline} built from an undergraduate final-year project (TFG) and a subsequent replication effort. Rather than proposing a novel method, we focus on a transparent, easy-to-reproduce setup that combines textual and audio cues and reports a first reference point on the SemEval-2024 Task 3 dataset.

Emotion recognition plays a critical role in fostering human connections and understanding. Emotions are deeply intertwined with human interactions, influencing communication, relationships, decision-making, and even societal structures \cite{Poria2017}. As digital interfaces become increasingly integral to our daily lives, the ability of machines to comprehend and appropriately respond to human emotions has garnered significant interest from researchers and industry professionals alike \cite{Picard2000}. Advances in Artificial Intelligence (AI), particularly in Natural Language Processing (NLP) and audio analysis, have opened promising avenues for automating emotion recognition \cite{Cambria2017}.

Traditional approaches to emotion recognition primarily rely on unimodal data sources, such as textual or audio inputs \cite{ElAyadi2011}. These methods, while effective to some degree, often fail to capture the nuanced interplay between verbal and non-verbal cues that humans use to express emotions \cite{Poria2019}. For instance, a speaker’s words might contradict their vocal tone, or subtle shifts in speech patterns might convey sarcasm, irony, or suppressed emotions. Such complexities cannot be fully understood without analyzing multiple modalities simultaneously. Recent research, including our study, seeks to address these limitations by adopting a multimodal approach, integrating textual and audio data to predict emotional effects in conversational contexts \cite{Zadeh2017}. Leveraging the SemEval dataset—a corpus annotated with cause-emotion pairs derived from the television series \textit{Friends}\cite{SemEval2024}—this study demonstrates how ensemble models can significantly enhance the accuracy and overall effectiveness of emotion recognition systems.

The integration of emotion recognition into AI systems has transformative potential across a range of fields. For instance, in mental health, AI systems can assist therapists by identifying early signs of emotional distress in patients through vocal tone or textual patterns in their speech \cite{Tzirakis2017}. In education, emotion-aware systems can adaptively modify teaching strategies based on real-time assessments of student engagement, providing personalized learning experiences \cite{Woolf2009}. Similarly, customer service platforms can use emotion recognition to provide empathetic and context-aware responses, improving user satisfaction and retention rates \cite{Seng2018}. Beyond these domains, emotion recognition systems have applications in languages, sociology, psychology, computer science, and even physiology \cite{Afzal2024}.

Despite its vast potential, current emotion recognition systems face critical limitations. Most notably, they often treat emotions as static, context-agnostic labels, neglecting the dynamic and context-dependent nature of human affect \cite{Cambria2017}. Emotions are rarely isolated states; they evolve over time and are influenced by cultural, situational, and interpersonal factors \cite{Schuller2013}. This makes multimodal approaches particularly promising, as they aim to emulate human-like emotional interpretation by synthesizing information from multiple data streams \cite{Poria2017}. By combining textual analysis with audio cues, the proposed research bridges this gap, enhancing the interpretability and accuracy of emotional signal detection in AI systems.

The SemEval dataset used in this study provides an ideal testbed for this approach. Its unique annotations of cause-emotion pairs allow for contextual emotion modeling, offering insights into how specific conversational elements lead to emotional reactions. This contextual framing is critical for applications requiring a nuanced understanding of emotional triggers and responses, such as in therapeutic dialogues or high-stakes customer interactions.

\subsection{Objectives}
\begin{itemize}
    \item To implement and evaluate simple, off-the-shelf transformer baselines for text and audio emotion recognition in conversations.
    \item To document a straightforward late-fusion ensemble and report its empirical behavior relative to unimodal baselines under a lightweight experimental protocol.
    \item To clearly state limitations (evaluation scope, tuning budget, and comparability) so results are interpreted as a reference point rather than a definitive benchmark.
\end{itemize}

\section{Theoretical Framework}
\subsection{Natural Language Processing (NLP)}
Natural Language Processing (NLP) focuses on enabling machines to process, analyze, and understand human language in both written and spoken forms. As a subfield of Artificial Intelligence, NLP bridges the gap between human communication and machine understanding, making it a cornerstone of many modern AI systems \cite{Jurafsky2000}. Key applications of NLP include machine translation, sentiment analysis, question-answering systems, text summarization, and notably, emotion recognition \cite{Mohammad2017}.

\subsubsection{Emotion Recognition in NLP}
Emotion recognition within NLP involves identifying and classifying affective states, such as happiness, sadness, anger, or fear, from textual data \cite{Canales2014}. This task goes beyond sentiment analysis by aiming to understand the underlying emotional tone of a message rather than merely categorizing it as positive, negative, or neutral. For instance, sentiment analysis might classify a text as negative, but emotion recognition would further specify whether the emotion conveyed is sadness, anger, or disgust.

NLP-based emotion recognition is a challenging task due to the inherent complexity of human language. Emotions are often conveyed indirectly through figurative language, metaphors, or sarcasm, which require advanced contextual understanding \cite{Seyeditabari2018}. Additionally, cultural and linguistic differences can significantly affect the way emotions are expressed, making cross-lingual emotion recognition another active area of research \cite{Balahur2012}.

\subsubsection{Role of Transformer Architectures}
Transformer-based architectures have revolutionized NLP by introducing attention mechanisms, allowing models to capture long-range dependencies and contextual relationships in text \cite{Vaswani2017}. Among these, BERT (Bidirectional Encoder Representations from Transformers) has set a new standard by pre-training on massive corpora and enabling bidirectional understanding of sentences. Unlike traditional NLP models, BERT can consider the full context of a word by looking at both the words preceding and succeeding it.

In the domain of emotion recognition, BERT has shown remarkable performance by accurately identifying subtle emotional cues in text \cite{Yang2021}. For example, BERT's ability to handle polysemy (words with multiple meanings) and ambiguous phrases enables it to distinguish between emotional states like sarcasm and genuine sadness.

\subsubsection{Variants of BERT for Emotion Recognition}
Several variants of BERT have been developed to address specific challenges or optimize performance in specialized tasks, including emotion recognition:
\begin{itemize}
    \item \textbf{RoBERTa (Robustly Optimized BERT)}: This variant improves upon BERT by removing constraints during training, such as next-sentence prediction, and utilizing larger datasets \cite{Liu2019}. These enhancements result in richer embeddings, which are particularly effective in capturing nuanced emotional tones in text.
     \item \textbf{DeBERTa (Decoding-enhanced BERT with disentangled attention)}: DeBERTa introduces a disentangled attention mechanism and an improved positional encoding method, enhancing its ability to capture syntactic and semantic relationships for emotion recognition tasks \cite{He2020}.
    \item \textbf{DistilRoBERTa}: A distilled version of RoBERTa, this variant provides a balance between computational efficiency and performance, making it well-suited for scenarios requiring faster inference times, such as real-time emotion detection \cite{Sanh2019}.
    \item \textbf{DistilBERT}: Designed as a lightweight version of BERT, DistilBERT retains much of BERT's effectiveness while reducing computational overhead. This makes it suitable for real-time emotion recognition applications, such as chatbots or virtual assistants, where efficiency is critical \cite{Sanh2019}.
\end{itemize}

\subsubsection{Advances and Challenges}
Recent advancements in NLP have introduced pre-trained language models capable of fine-tuning for emotion recognition tasks. For instance, transfering learning from related tasks in speech and text to adapt to unseen emotions and domains, leveraging shared knowledge to improve accuracy \cite{Ananthram2020}. Moreover, large-scale emotion datasets, such as EmoReact \cite{Nojavanasghari2016} and ISEAR \cite{Scherer2013}, provide the necessary diversity to train and evaluate these models comprehensively.

However, challenges remain in fully leveraging NLP for emotion recognition:
\begin{itemize}
   \item \textbf{Subtlety of Emotions}: Emotions like anticipation or surprise are often inferred from context rather than explicitly stated. Similarly, tasks like moral value detection share this subtlety \cite{Yeste2024}, requiring models to interpret implied or context-dependent cues, such as ethical dilemmas or implicit societal norms embedded in text. Models must develop a deeper understanding of implied meaning to excel in these nuanced domains \cite{Canales2014}.
    \item \textbf{Code-Switching}: In multilingual contexts, individuals frequently switch between languages. Capturing emotions accurately in such scenarios requires models capable of understanding multiple languages and their interplay \cite{Balahur2012}.
    \item \textbf{Figurative Language}: Idioms, metaphors, and sarcasm pose significant challenges. For example, the phrase "I'm over the moon" conveys happiness, but its literal interpretation would confuse a less sophisticated model \cite{Strapparava2007}.
    \item \textbf{Dataset Bias}: Many emotion datasets lack diversity in cultural representation, leading to biased models that underperform on texts from underrepresented groups \cite{Teye2022}.
\end{itemize}

\subsubsection{Future Directions}
To further advance NLP-driven emotion recognition, researchers are exploring multimodal systems that integrate text with additional data sources, such as audio and visual cues. For instance, analyzing textual data alongside vocal intonations and facial expressions can provide a holistic view of emotional states \cite{Majumder2018}. Moreover, incorporating knowledge graphs and external world knowledge into NLP models holds promise for improving contextual understanding, especially in complex conversational scenarios.

Applications of emotion recognition in NLP continue to grow, including mental health support systems, emotion-aware customer service bots, and interactive storytelling. These applications highlight the potential of NLP to not only understand but also respond empathetically to human emotions, paving the way for more natural and meaningful human-machine interactions.

\subsection{Audio Analysis with AI}
Speech conveys a wealth of emotional information through variations in tone, pitch, rhythm, volume, and speech rate. These vocal attributes are crucial in conveying affective states, often complementing or even contradicting the semantic content of spoken words \cite{Cowen2017}. Emotion recognition through speech analysis has been a significant focus in affective computing, as it enables systems to infer emotions in real-time applications, such as virtual assistants, call centers, and mental health monitoring.

\subsubsection{Traditional Approaches}
Traditional audio processing techniques have relied on manually engineered features, such as Mel-frequency cepstral coefficients (MFCC), pitch contours, and formant frequencies \cite{Eyben2010}. These features represent the spectral and prosodic properties of speech and are typically used as inputs for machine learning classifiers like support vector machines (SVMs) and hidden Markov models (HMMs). While effective, these approaches are constrained by the limited ability of hand-crafted features to capture the nuances of vocal emotions \cite{Ververidis2006}.

\subsubsection{Modern Advances with Deep Learning}
Modern AI models have revolutionized audio analysis by learning feature representations directly from raw audio data. Wav2vec 2.0, a state-of-the-art self-supervised model, exemplifies this paradigm \cite{Baevski2020}. Unlike traditional methods, wav2vec 2.0 utilizes convolutional neural networks (CNNs) to encode audio waveforms into compact feature representations, followed by transformers to capture temporal dependencies across speech segments. This architecture enables the model to generalize effectively across diverse datasets and scenarios without requiring extensive labeled data.

In emotion recognition tasks, wav2vec 2.0 excels in capturing subtle vocal cues, such as stress, intonation, and variations in speech rhythm, which are often overlooked by text-only systems \cite{caulley2023} \cite{lesyk2024}. This capability stems from its self-supervised learning approach, which involves masking segments of raw audio and reconstructing them, thereby generating robust, contextualized representations \cite{wang2025speech}. These cues provide critical insights into affective states, making wav2vec 2.0 an ideal component in multimodal emotion recognition frameworks.

\subsubsection{Applications of Audio-Based Emotion Recognition}
Speech-based emotion recognition has found applications in several domains:
\begin{itemize}
    \item \textbf{Mental Health Monitoring}: Systems equipped with speech emotion analysis can detect signs of stress, anxiety, or depression, offering early intervention opportunities \cite{Cummins2015}.
    \item \textbf{Customer Service}: Call center analytics use vocal emotion recognition to assess customer satisfaction and agent performance in real-time interactions \cite{Parra2022}.
    \item \textbf{Education}: Speech-based systems can analyze student emotions to provide adaptive feedback and higher education management \cite{Zhou2022}.
    \item \textbf{Entertainment and Gaming}: Emotion-aware voice assistants and characters create immersive and interactive user experiences \cite{Ringeval2013}.
\end{itemize}

\subsubsection{Future Directions}
Despite advancements, several challenges remain in audio-based emotion recognition:
\begin{itemize}
    \item \textbf{Environmental Noise}: Background noise and overlapping speech reduce the accuracy of emotion detection in real-world settings \cite{Schuller2018}.
    \item \textbf{Cultural and Linguistic Variations}: Emotion expression varies significantly across cultures and languages, requiring models to adapt effectively to diverse populations \cite{Cortiz2022}.
    \item \textbf{Dataset Limitations}: Many existing speech emotion datasets lack diversity in terms of demographics and emotional range, leading to biases in model performance \cite{Dair2021}.
\end{itemize}

Future research aims to address these challenges by leveraging multimodal data, such as combining audio with visual and physiological signals, and developing noise-robust models capable of real-time emotion analysis. Additionally, advances in self-supervised learning and domain adaptation will enable models to generalize better across languages and environments.

\subsection{Multimodal Approaches}
Multimodal learning integrates information from multiple sources, such as text, audio, and visual data, to provide a holistic understanding of complex phenomena. In the context of emotion recognition, multimodal approaches are particularly effective as they capture both semantic and non-verbal cues, such as textual meaning, vocal tone, facial expressions, and physiological signals \cite{Zadeh2017,Kessous2010}. These approaches address the limitations of unimodal systems, which often fail to interpret emotions conveyed through subtle and context-dependent signals.

\subsubsection{Fusion Strategies in Multimodal Learning}
To combine information from different modalities, various fusion strategies are employed:
\begin{itemize}
    \item \textbf{Early Fusion}: Features from different modalities are concatenated at the input level and fed into a single model for joint learning. While this method captures low-level interactions, it requires aligned data and can suffer from high dimensionality \cite{Baltrusaitis2018}.
    \item \textbf{Late Fusion}: Each modality is processed independently, and predictions are combined at the decision level using ensemble techniques such as weighted averaging or voting. Late fusion improves system robustness by leveraging modality-specific strengths \cite{Snoek2005}.
    \item \textbf{Hybrid Fusion}: Combines elements of both early and late fusion, allowing interactions at intermediate stages of model processing \cite{Poria2017}.
\end{itemize}

\subsubsection{Applications of Multimodal Emotion Recognition}
Multimodal emotion recognition has seen applications across a variety of fields:
\begin{itemize}
    \item \textbf{Healthcare}: Multimodal systems combine speech analysis with physiological signals, such as heart rate and galvanic skin response, to detect stress, depression, and other mental health conditions \cite{Cummins2015}.
    \item \textbf{Human-Computer Interaction (HCI)}: Multimodal interfaces incorporating text, speech, and vision enable more natural and empathetic interactions between users and machines \cite{Vinciarelli2009}.
    \item \textbf{Entertainment}: Emotion-aware systems in gaming and virtual reality integrate facial expressions, voice intonation, and body gestures to create immersive experiences \cite{Ringeval2013}.
\end{itemize}

\subsubsection{Challenges}
Despite their advantages, multimodal approaches face several challenges:
\begin{itemize}
    \item \textbf{Data Alignment}: Ensuring temporal alignment across modalities, especially for speech and video, is computationally intensive and prone to errors \cite{Schmitz2022}.
    \item \textbf{Modality Noise and Missing Data}: Noisy or missing data in one modality can negatively impact the overall system performance. Robust fusion techniques are required to mitigate this issue \cite{Morency2017}.
    \item \textbf{Computational Complexity}: Multimodal models are resource-intensive, requiring significant computational power for training and inference \cite{Jin2024}.
\end{itemize}

\subsubsection{The Proposed Ensemble Strategy}
This study employs an ensemble strategy that combines textual and audio-based models. The textual component leverages advanced NLP techniques to extract semantic meaning, while the audio component captures paralinguistic features such as tone and pitch. By leveraging the complementary strengths of these modalities, the ensemble model achieves enhanced accuracy and robustness \cite{Tzirakis2017}.

\section{Methodology}
\subsection{Dataset and Preprocessing}
The dataset used in this study is sourced from the television series \textit{Friends}, which contains dialogues annotated with cause-emotion pairs and has been used as a task at SemEval2024 \cite{SemEval2024}. Each annotation links a conversational utterance to the emotional effects it elicits, providing both textual and auditory data for analysis. This dataset is particularly valuable as it reflects realistic conversational dynamics and emotional exchanges.

Preprocessing involved the following steps:
\begin{itemize}
    \item Extracting and aligning textual and audio data to ensure each conversational utterance had a corresponding representation in both modalities.
    \item Standardizing audio signals to a uniform sampling rate of 16kHz for consistency in processing and compatibility with the audio analysis models.
    \item Structuring the dataset into training and test splits, ensuring balanced representation across emotional categories.
\end{itemize}

These preprocessing steps ensured the dataset was well-prepared for multimodal emotion recognition tasks, facilitating seamless integration into the experimental pipeline.

\subsection{Model Architecture}
The model architecture was designed to leverage both textual and audio data for multimodal emotion recognition. Two primary components were implemented to process these modalities:

\begin{itemize}
    \item \textbf{Text Models:} RoBERTa, DistilBERT, DeBERTa, and DistilRoBERTa transformer-based models were fine-tuned to classify emotional states from the textual data. These models utilized pre-trained embeddings that were further optimized during training on the dataset to capture the semantic and contextual nuances of conversational text.
    \item \textbf{Audio Models:} HuBERT, Wav2Vec2, and Wav2Vec2-large-robust were employed for processing audio clips. An architecture such as Wav2Vec2 is particularly effective in encoding both temporal and spectral features of speech, enabling the extraction of paralinguistic information, such as tone and intonation, relevant to emotion recognition.
\end{itemize}

An ensemble model was constructed to integrate predictions from both modalities. By combining textual and audio outputs, the ensemble approach enhanced the robustness and accuracy of emotion classification, ensuring a comprehensive understanding of the emotional states expressed in the data.

\subsection{Training and Evaluation}
The models were trained using an optimizer configured to ensure stable convergence during the learning process. The dataset was divided into training, and test splits, with proportions set to 80\% for training, and 20\% for testing.

Key evaluation metrics included:
\begin{itemize}
    \item \textbf{Accuracy:} Measuring the proportion of correctly identified emotional states among the predictions.
    \item \textbf{Execution Time:} Evaluating the computational performance of the models during both training and inference phases.
\end{itemize}

The evaluation process involved assessing the models on the test set after training to ensure the approach generalized well to unseen data. The results highlighted the effectiveness of the multimodal ensemble strategy in improving classification performance compared to individual modalities.

\subsection{Results and Discussion}

\subsubsection{Text Models}
The performance of text models was evaluated using a subset of the dataset to maintain comparability with audio models. Four transformer-based models were considered: RoBERTa, DistilBERT, DeBERTa, and DistilRoBERTa-sentiment-finetunned (DistilRoBERTa).

\textbf{Accuracy Results:}
The models demonstrated varying levels of accuracy when predicting emotional states:
\begin{itemize}
    \item \textbf{RoBERTa:} 50.68\%
    \item \textbf{DistilBERT:} 49.82\%
    \item \textbf{DeBERTa:} 48.29\%
    \item \textbf{DistilRoBERTa:} 41.83\%
\end{itemize}
RoBERTa achieved the highest accuracy among the text models, likely due to its optimized training and ability to capture contextual dependencies effectively. DistilRoBERTa, while fine-tuned for sentiment analysis, underperformed in this task, possibly due to domain-specific biases.

\textbf{Execution Time:}
The execution times were as follows:
\begin{itemize}
    \item \textbf{RoBERTa:} 74.6 seconds
    \item \textbf{DistilBERT:} 51.25 seconds
    \item \textbf{DeBERTa:} 21.56 seconds
    \item \textbf{DistilRoBERTa:} 5.61 seconds
\end{itemize}
DistilRoBERTa was the fastest, reflecting its lightweight architecture. However, its lower precision underscores the trade-off between speed and accuracy.

\subsubsection{Audio Models}
Three audio models were fine-tuned and evaluated for emotion recognition: HuBERT-ls960, Wav2Vec2, and Wav2Vec2-large-robust-12-ft-emotion-msp-dim.

\textbf{Accuracy Results:}
The audio models achieved the following accuracy:
\begin{itemize}
    \item \textbf{HuBERT:} 31.08\%
    \item \textbf{Wav2Vec2:} 35.43\%
    \item \textbf{Wav2Vec2-large-robust:} 32.54\%
\end{itemize}
Wav2Vec2 outperformed the other audio models, likely benefiting from its robust feature extraction capabilities. The relatively lower accuracy compared to text models highlights the challenges of emotion recognition based solely on vocal cues.

\textbf{Execution Time:}
The execution times were as follows:
\begin{itemize}
    \item \textbf{HuBERT:} 113.6 seconds
    \item \textbf{Wav2Vec2:} 87.25 seconds
    \item \textbf{Wav2Vec2-large-robust:} 97.32 seconds
\end{itemize}
HuBERT exhibited the longest execution time, reflecting its computationally intensive processing. Wav2Vec2's balance of performance and efficiency reinforces its suitability for this task.

\subsubsection{Ensemble Model}
An ensemble model combining Wav2Vec2 (audio) and RoBERTa (text) was evaluated to leverage the strengths of both modalities.

\textbf{Accuracy Result:}
The ensemble model achieved an accuracy of \textbf{62.97\%}, significantly outperforming individual text and audio models. This improvement underscores the advantages of multimodal integration, as the combination of textual and audio features provides a more comprehensive understanding of emotional expressions.
\section{Limitations and Scope}
\label{sec:limitations}
This manuscript should be interpreted as a baseline technical report. Key limitations include:
\begin{itemize}[leftmargin=*]
    \item \textbf{Lightweight evaluation:} we primarily report accuracy and wall-clock execution time; class-imbalance-aware metrics (e.g., Macro-F1) and statistical stability analyses (multiple random seeds) are left for future work.
    \item \textbf{Limited tuning budget:} hyperparameter search and extensive ablations were not performed; results may change under stronger tuning.
    \item \textbf{Comparability to SemEval systems:} our splits and training protocol may differ from official setups; therefore, results should not be treated as directly comparable to leaderboard submissions.
    \item \textbf{Dataset and domain constraints:} the dataset is derived from \textit{Friends} and may not generalize to other domains, languages, or interaction styles.
    \item \textbf{Modality coverage:} we do not use visual information; missing/noisy audio segments can reduce the contribution of the speech modality.
\end{itemize}

\subsubsection{Discussion}
The results highlight the superiority of multimodal approaches over unimodal models. While text models generally exhibited higher accuracy than audio models, the ensemble approach demonstrated the potential to synthesize complementary modalities for enhanced performance. RoBERTa's consistent outperformance among text models and Wav2Vec2's robustness in audio classification validate their selection for the ensemble. However, the trade-off between computational efficiency and accuracy warrants consideration, particularly for real-time applications.

\section{Conclusions}

This study demonstrates the effectiveness of a multimodal approach to emotion recognition, integrating textual and audio data to achieve higher accuracy compared to unimodal methods. Leveraging the SemEval 2024 dataset, the analysis explored state-of-the-art transformer models for text (e.g., RoBERTa, DistilBERT, DeBERTa) and audio (e.g., Wav2Vec2, HuBERT), as well as an ensemble strategy that combined the strengths of both modalities.

The results highlight that text models consistently outperformed audio models in terms of accuracy, with RoBERTa achieving the highest accuracy among textual approaches. Audio models, while providing valuable paralinguistic cues, demonstrated limitations in capturing nuanced emotional states on their own. The ensemble model, combining Wav2Vec2 and RoBERTa, achieved the highest overall accuracy of 62.97\%, underscoring the potential of multimodal integration to enhance emotion recognition.

This work contributes to the ongoing advancement of affective computing, offering insights into how machines can better understand and respond to human emotions in complex, real-world interactions. While the findings affirm the advantages of multimodal systems, they also reveal challenges, including increased computational demands and the need for balanced, well-aligned datasets. Future research should explore the inclusion of additional modalities, such as facial expressions, physiological signals, or contextual metadata, and develop more efficient methods for real-time application. Addressing dataset biases and extending the scope to cross-cultural and multilingual scenarios will further enhance the robustness and applicability of multimodal emotion recognition systems.

%%
%% Define the bibliography file to be used
\printbibliography

\end{document}